\documentclass[lettersize,journal]{IEEEtran}
\usepackage{amsmath,amsfonts}
\usepackage{algorithmic}
\usepackage{algorithm}
\usepackage{array}
\usepackage{textcomp}
\usepackage{stfloats}
\usepackage{url}
\usepackage{verbatim}
\usepackage{graphicx}
\usepackage{cite}
\usepackage[colorlinks,
            linkcolor=blue,
            anchorcolor=green,
            citecolor=red
            ]{hyperref}
\usepackage{tikz}
\usepackage{comment}
\usepackage{color}
\usepackage{rotating}
\usepackage{subfigure}
\usepackage{multirow} 
\usepackage{caption}
\usepackage{subcaption}
\usepackage{amssymb}
\definecolor{nbarrier}{RGB}{255, 150, 255}
\definecolor{nbicycle}{RGB}{100, 230, 245}
\definecolor{nbus}{RGB}{0, 0, 255}
\definecolor{ncar}{RGB}{100, 150, 245}
\definecolor{nconstruct}{RGB}{150, 30,90}
\definecolor{nmotor}{RGB}{30, 60, 150}
\definecolor{npedestrian}{RGB}{250, 30, 30}
\definecolor{ntraffic}{RGB}{255, 92, 0}
\definecolor{ntrailer}{RGB}{255, 40, 200}
\definecolor{ntruck}{RGB}{80, 30, 180}
\definecolor{ndriveable}{RGB}{255, 0, 255}
\definecolor{nother}{RGB}{175, 0, 75}
\definecolor{nsidewalk}{RGB}{75, 0, 75}
\definecolor{nterrain}{RGB}{150, 240, 80}
\definecolor{nmanmade}{RGB}{255, 200, 0}
\definecolor{nvegetation}{RGB}{0, 175, 0}
\definecolor{nothers}{RGB}{0, 0, 0}

\definecolor{kcar}{RGB}{99, 149, 244}
\definecolor{kbicycle}{RGB}{99, 229, 244}
\definecolor{kmotorcycle}{RGB}{30, 60, 149}
\definecolor{ktruck}{RGB}{79, 30, 179}
\definecolor{kothervehicle}{RGB}{0, 0, 255}
\definecolor{kperson}{RGB}{255, 30, 30}
\definecolor{kbicyclist}{RGB}{255, 40, 200}
\definecolor{kmotorcyclist}{RGB}{149, 30, 90}
\definecolor{kroad}{RGB}{255, 0, 255}
\definecolor{kparking}{RGB}{255, 149, 255}
\definecolor{ksidewalk}{RGB}{75, 0, 75}
\definecolor{kotherground}{RGB}{175, 0, 75}
\definecolor{kbuilding}{RGB}{255, 200, 0}
\definecolor{kfence}{RGB}{255, 120, 49}
\definecolor{kvegetation}{RGB}{0, 175, 0}
\definecolor{ktrunk}{RGB}{134, 60, 0}
\definecolor{kterrain}{RGB}{149, 239, 79}
\definecolor{kpole}{RGB}{255, 239, 149}
\definecolor{ktrafficsign}{RGB}{255, 0, 0}

\begin{document}

\title{Multimodal Point Cloud Semantic Segmentation With Virtual Point Enhancement}



\author{Zaipeng Duan$^{1,2}$, Xuzhong Hu$^{1,2}$,Pei An$^{3}$, Jie Ma$^{1,2,*}$
\thanks{$^{1}$School of Artificial Intelligence and Automation, Huazhong University of Science and Technology (HUST), Wuhan, Hubei 430074, China}
\thanks{$^{2}$National Key Laboratory of Science and Technology on Multispectral Information Processing, HUST, Wuhan, Hubei 430074, China.}
\thanks{$^{3}$School of Electronic Information and Communications,
Huazhong University of Science and Technology, Wuhan, China.}}



\maketitle

\begin{abstract}
LiDAR-based 3D point cloud recognition has been proven beneficial in various applications. However, the sparsity and varying density pose a significant challenge in capturing intricate details of objects, particularly for medium-range and small targets. Therefore, we propose a multi-modal point cloud semantic segmentation method based on Virtual Point Enhancement (VPE), which integrates virtual points generated from images to address these issues. These virtual points are dense but noisy, and directly incorporating them can increase computational burden and degrade performance. Therefore, we introduce a spatial difference-driven adaptive filtering module that selectively extracts valuable pseudo points from these virtual points based on density and distance, enhancing the density of medium-range targets. Subsequently, we propose a noise-robust sparse feature encoder that incorporates noise-robust feature extraction and fine-grained feature enhancement. Noise-robust feature extraction exploits the 2D image space to reduce the impact of noisy points, while fine-grained feature enhancement boosts sparse geometric features through inner-voxel neighborhood point aggregation and downsampled voxel aggregation. The results on the SemanticKITTI and nuScenes, two large-scale benchmark data sets, have validated effectiveness, significantly improving 2.89\% mIoU with the introduction of 7.7\% virtual points on nuScenes.
\end{abstract}


\begin{IEEEkeywords}
Semantic segmentation, Multi-modal, Deep learning, Point clouds.
\end{IEEEkeywords}

\maketitle

\section{Introduction}
Scene understanding is crucial for autonomous driving and robotics \cite{1,72}. As a pivotal task within scene understanding, point cloud semantic segmentation aims to predict category labels for each point in the LiDAR point cloud. In recent years, the academic community has focused on exploring the use of camera images \cite{4} or LiDAR point clouds \cite{7,9} to understand natural environments. However, a single modality struggles to ensure robust perception in diverse scenarios. Concretely, cameras offer rich texture and color information but lack depth information and are sensitive to lighting variation. In contrast, LiDAR can capture sparse point data while providing accurate depth information. The complementary information from LiDAR and cameras can enhance scene understanding.

\begin{figure}[t]
\centering
\includegraphics[width=0.45\columnwidth]{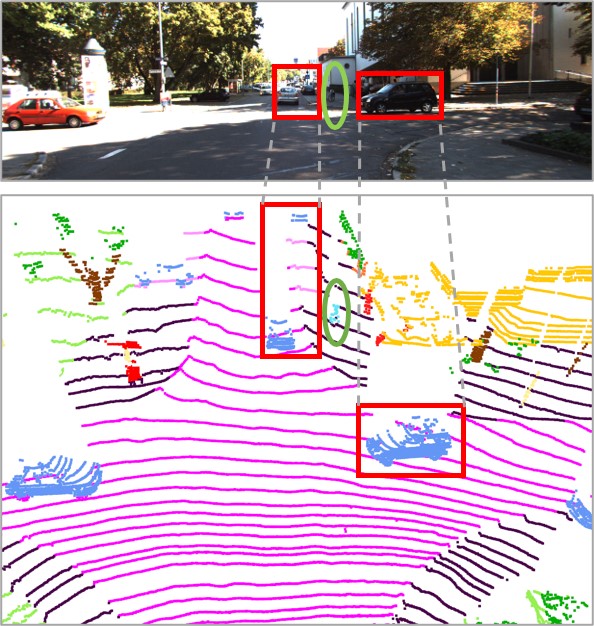}
\includegraphics[width=0.45\columnwidth]{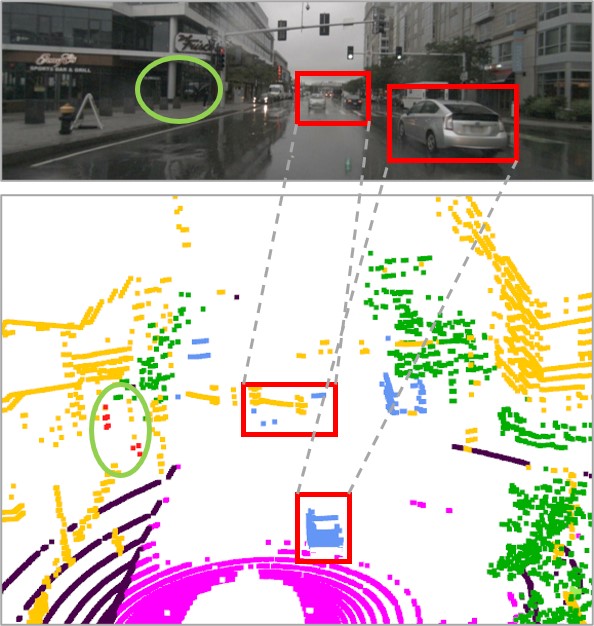}
\caption{ The sparsity of point clouds. The left case is from the SemanticKITTI dataset and the right case is from the NuScenes dataset. As distance increases, LiDAR points become sparser, making it challenging to identify medium-range targets such as cars (marked with red boxes) and small targets with only a few points, such as pedestrians and bicycles (marked with green circles).}
\label{0}
\end{figure}

Recently, both cameras and LiDAR have been used on many commercially produced vehicles, which has spurred academic research into multimodal data fusion \cite{11,12,13}. These methods can be broadly categorized into two primary categories: The first projects the point cloud onto the image plane using the extrinsic between sensors and the camera's intrinsic to obtain a projected depth map. This is then fused with the corresponding image features through a 2D Convolutional Neural Network (CNN) to obtain segmentation results in the co-visible region. The other acquires point cloud features and image features separately through 3D CNN and 2D CNN, respectively. Based on the mapping relationship between points and pixels, co-visible pixel features are fused into the point features, resulting in point-wise scores. However, these fusion-based approaches still have unavoidable limitations: despite the introduction of camera texture information, it still hasn't altered the inherent sparsity of the point cloud, which limits medium-range and small targets (see Fig. \ref{0}). 

To address the aforementioned challenges, we draw inspiration from the successful application of virtual points \cite{15,16} in 3D object detection and propose a virtual point enhanced multi-modal semantic segmentation method.  Specifically, it focuses on selecting high-quality pseudo points from virtual points generated by image depth completion or instance segmentation seeds and utilizes a noise-robust encoder to extract fine-grained features. First, we introduce a spatial difference-driven adaptive filtering module based on distance and density, which aims to extract reliable pseudo points from dense but noisy virtual points, effectively enhancing the density of medium-range targets. Second, we propose a noise-robust sparse feature encoder that incorporates noise-robust feature extraction and fine-grained feature enhancement. Noise-robust feature extraction reduces noise disturbances by encoding geometric features in both 2D image and 3D LiDAR spaces, while fine-grained feature enhancement improves feature representation through  inner-voxel neighborhood point aggregation and increases the receptive field through downsampled voxel aggregation, thereby enhancing sparse geometric features. Experimental results on the SemanticKITTI \cite{17} dataset and the nuScenes \cite{18} dataset validate the effectiveness of VPENet. 

In general, our contributions are as follows:

1) We propose a multi-modal point cloud semantic segmentation method based on virtual point enhancement, named the VPENet. It effectively alleviates point cloud sparsity by introducing virtual points generated from images and applying a spatial difference-driven adaptive filtering module based on distance and density to selectively retain reliable virtual points, notably improving 2.89\% mIoU with the introduction of 7.7\% virtual points on nuScenes.

2) We developed a noise-robust sparse feature encoder that incorporates noise-robust feature extraction and fine-grained feature enhancement. It reduces the impact of noisy points by leveraging the 2D image space and enhances sparse fine-grained features by combining intra-voxel point feature aggregation with downsampled voxel feature aggregation, without the additional feature decoder.

3) Our method achieves competitive results across multiple LiDAR semantic segmentation benchmarks, offering a solution for integrating virtual points in LiDAR point cloud segmentation.

\section{RELATED WORK}
\subsection{Single-Model Methods}
Camera-based semantic segmentation aims to predict pixel-wise labels for 2D images. FCN \cite{19} is the first end-to-end fully convolutional network for image semantic segmentation. Beyond FCN, recent research has made significant progress by exploring methods such as multi-scale information \cite{21}, atrous convolution \cite{4,23}, attention mechanisms \cite{24,25}, and Transformer \cite{26,27}. However, methods using only cameras face uncertainties in depth perception and are sensitive to lighting interference. LiDAR-based methods process point clouds using several primary representations. 1) Point-based methods operate directly on points with PointNet \cite{29} being the first major contribution in this field. Subsequently, point-based MLPs \cite{1,30}, point convolution \cite{31,32} methods for neighborhood feature extraction, and Transformer \cite{63,64} for long-range dependencies were proposed. However, point-based methods become computationally intensive as the number of points increases, making them unsuitable for complex outdoor scenes. 2) Projection-based methods efficiently handle LiDAR point clouds by projecting them onto 2D pixels, enabling the use of traditional CNNs. Previous studies have converted LiDAR points into 2D images using the bird's eye view (BEV) projection \cite{33,34}, spherical projection \cite{7,35,36}, or both \cite{38}. However, this projection inevitably results in information loss. 3) Voxel-based methods achieve a balance between efficiency and effectiveness with the introduction of sparse convolution \cite{39}. SparseConv improves efficiency over traditional voxel-based methods (e.g., 3DCNNs) by storing only nonempty voxels in a hash table and performing convolutions only on them. Cylinder3D \cite{40} transforms the original voxel representation into cylindrical voxels and designs an asymmetric residual block to boost performance. AF2-S3Net \cite{41} uses multi-branch kernels of varying sizes and an attention mechanism to aggregate multi-scale features, improving the segmentation of smaller objects. OA-CNNS \cite{65} adaptively adjusts receptive fields on non-overlapping pyramid grids, effectively perceiving contextual information. However, these methods rely solely on sparse, textureless LiDAR point clouds, neglecting the appearance and texture information of images.

\subsection{Multi-Model Fusion Methods}
Multi-sensor fusion methods integrate complementary information from cameras and LiDAR, leveraging the strengths of both. RGBAL \cite{11} converts RGB images into polar coordinate network mappings and has developed strategies for early data fusion and mid-stage feature fusion. PointPainting \cite{12} projects images onto LiDAR space using BEV projection or spherical projection to improve the performance of the LiDAR network. PMF \cite{13} synergistically fuses the appearance information of the RGB images in the camera coordinates with the spatial depth information of point clouds. 2Dpass \cite{42} refines multi-modal information into a single point cloud modality, thereby reducing inference time. 2D3DNet \cite{44} uses a pre-trained 2D model to generate per-pixel semantic labels, reducing the reliance on 3D labeled data. Mseg3D \cite{43} alleviates the heterogeneity of modality by optimizing the extraction of single-modal features and the fusion of multimodal features. Lif-seg \cite{60} effectively addresses the weak spatiotemporal synchronization problem through early coarse fusion and fine fusion after modal alignment. However, these methods have not addressed the sparsity of point clouds that limits the segmentation of distant and small targets.

\subsection{Virtual Point In Object Detection Methods}
RGB images and LiDAR are complementary and often enhance 3D perception performance. Recent research has adopted virtual points to fuse these two types of data. MVP \cite{45} enhances point clouds by selecting seed points from foreground objects and projecting them into point cloud space. Inspired by MVP's seed sampling strategy, UVTR \cite{46} establishes LiDAR and camera branches in voxel space to preserve their respective specific information and achieve mutual interaction through voxel addition operations. SFD \cite{15} fuses the original point cloud features with the virtual points obtained from the image depth estimation in a more refined manner. MSMDF \cite{47} finely controls voxels through multi-depth de-shadowing and modality-aware gating. Virconv \cite{16} addresses noise interference in depth estimation with random voxel drop and anti-noise submanifold convolution. Although virtual points clearly capture the geometric shape of distant objects through depth estimation, demonstrating the prospect for high-powered 3D perception, they are often very dense and prone to noise. In this paper, we employ an adaptive filtering technique based on distance and density to selectively extract valuable pseudo points from the dense but noisy virtual points.
\begin{figure*}[t]
\centering\includegraphics[width=1.0\textwidth]{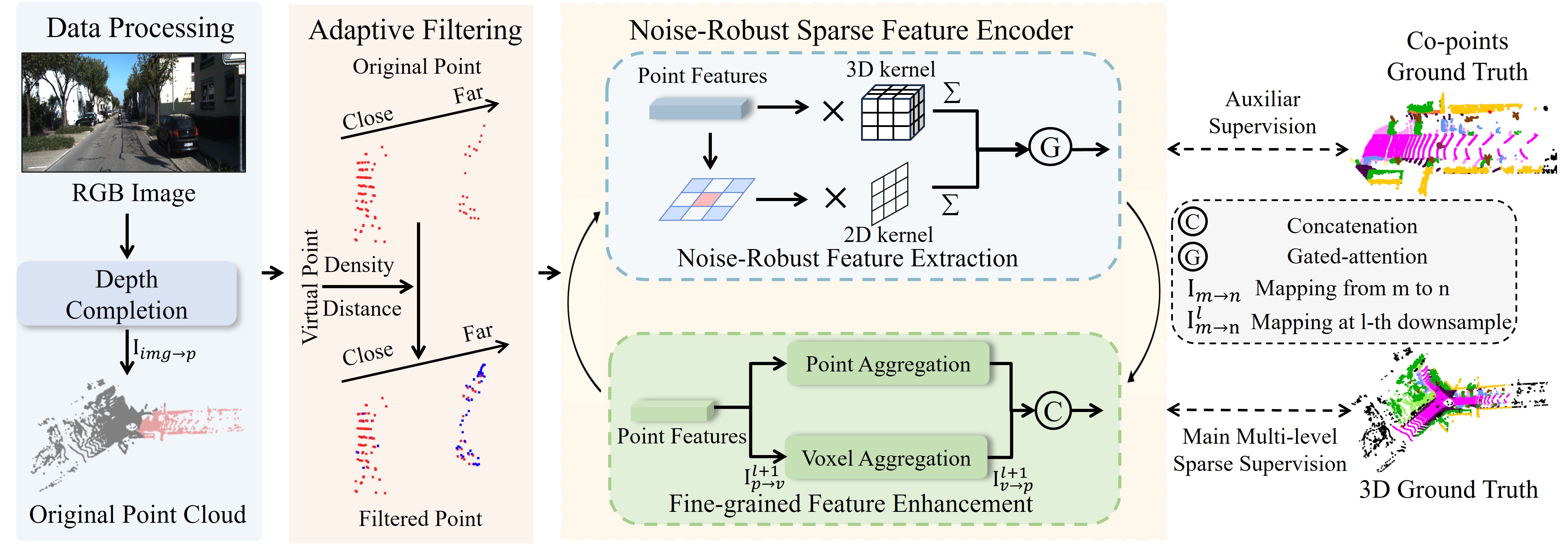}
\caption{Overview of our Virtual Point Enhancement (VPE) for Multi-modal Point Cloud Semantic Segmentation. It first through depth completion to obtain the dense but noisy virtual points from the original camera images. These points (in blue) are then merged with the original points (in red) and fed into the spatial difference-driven adaptive filtering module to increase the point cloud density of medium-range and small targets. Subsequently, the merged point cloud undergoes noise reduction through noise-robust feature extraction and enhancement with sparse geometric features, generating input for the next stage of the noise-robust sparse feature encoder. Finally, the co-visible point sparse supervision will be incorporated as an auxiliary loss to complement the primary multi-layer sparse supervision in the fine-grained feature enhancement.}
\label{net}
\end{figure*}
\section{METHOD}

The overall network structure includes three modules: 1) spatial difference-driven adaptive filtering module and 2) noise-robust sparse feature encoder that incorporates noise-robust feature extraction and fine-grained feature enhancement, as shown in Fig. \ref{net}. Specifically, we first generate numerous dense but noisy virtual points from RGB images by depth estimation. Then, we employ a spatial difference-driven adaptive filtering module to filter out relatively reliable pseudo points, mitigating the sparsity of the point cloud. Subsequently, we integrate the pseudo points with the original points, feeding them into the noise-robust sparse feature encoder. This module reduces the impact of noisy points through noise-robust feature extraction and enhances sparse geometric features through fine-grained feature enhancement, generating the input for the next stage of the sparse feature encoder.

\subsection{Prerequisite}
$Voxelization.$  Voxelization is the process of converting discrete point clouds $P=\left\{\left(x_{i}, y_{i}, z_{i}\right)\right\}_{i=1}^{N}$ into a regular grid based on a predefined grid scale $s_{l}$. The process is as follows:
\begin{equation}
V_{l}^{\text {voxel }}=\left\{\left(\left\lfloor x_{i} / s_{l}\right\rfloor,\left\lfloor y_{i} / s_{l}\right\rfloor,\left\lfloor z_{i} / s_{l}\right\rfloor\right)\right\}_{i=1}^{N} \in \mathbb{R}^{N \times 3}
\end{equation}
where $V_{l}^{\text {voxel }}$ is the voxel index in the $l$-th layer, $\lfloor$ is the ceiling function. Through this, we can use the nearest interpolation on the sparse voxel features to obtain point-wise 3D features. 
$Scatter$ $\Phi_{\mathcal{P} \rightarrow \mathcal{V}}^{s}$ and ${Gather}$ $\Phi_{\mathcal{V} \rightarrow \mathcal{P}}^{s}.$  A mapping framework between point $P$ and voxel $V$ coordinates for indexing is developed. The Scatter and Gather operations enable mutual conversion between voxel features $F^{s}_{v}$ and original point features $F^{s}_{p}$ at voxel scale $s$, where $F^{s}_{v} \in \mathbb{R}^{{N_{V}}\times{C}}$, $F^{s}_{p} \in \mathbb{R}^{N \times{C}}$ and $C$ is the number of channels. In the scatter operation at the voxel scale $s$, voxel features are obtained by aggregating point features with the same voxel index using the mean or max operations. The inverse operation, Gather, collects pointwise features from voxel features via indexing (i.e., copying).
$Point-to-Pixel$ $Correspondence.$  We can obtain the index between points and image pixels using the given extrinsic ${T}$ matrix between sensors and the camera's intrinsic ${K}$ matrix. Let $^{L}{P}_{N}\in \mathbb{R}^{4 \times{N}}$ be one of the training sets with ${N}$ points. Each point $^{L}{p}_{i}$ in $^{L}{P}_{N}$ includes 3D coordinates (${x}, {y}, {z}$) and a reflection intensity ($\mathrm{i}$), and the obtained 2D projection pixel coordinate $^{C}{p}_{i}=\left[\begin{array}{ll}u_{i} & v_{i}\end{array}\right] \in \mathbb{R}^{{H} \times{W}}$ and H and W denote the height and width of the image, respectively. To apply the coordinate transformation, we add a fourth column to $^{L}{p}_{i}$, turning it into a 4D vector, making the equation homogeneous. The projection process is described as follows:
\begin{equation}
\begin{aligned}
 ^{C}{p}_{i}& ={{K}} {T} {^{L}{\widetilde{p}}}_{i} /{Z}_{i} \\
& ={{K}} [{R} \mid {t}] \left[\begin{array}{llll} x_{i} &y_{i} &z_{i} &1 
\end{array}\right]^{T} / {Z}_{i}
\end{aligned}
\end{equation}
\begin{equation}
{T}=\left[\begin{array}{cc}
{R} & {t} \\
0 & 1
\end{array}\right]
\end{equation}
where $^{L} {\widetilde{p}}_{i}$ and $^{C}{p}_{i}$ are the corresponding 3D point coordinates and image pixel coordinates, and ${R}$ and ${t}$ represent the rotation matrix and the translation vector in the transformation matrix ${T}$, ${Z}_{i}$ is the depth of the 3D point.
\subsection{Spatial Difference-driven Adaptive Filtering Module}
\begin{figure*}[t]
\centering
\subfigure[Images]{
\includegraphics[width=0.31\textwidth]{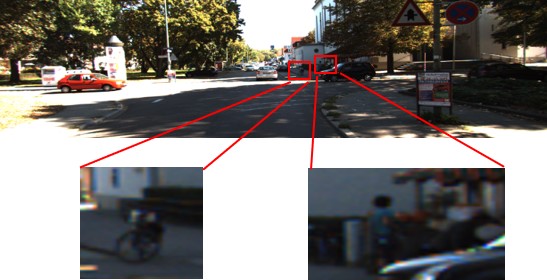}}
\subfigure[Depth Completion]{
\includegraphics[width=0.31\textwidth]{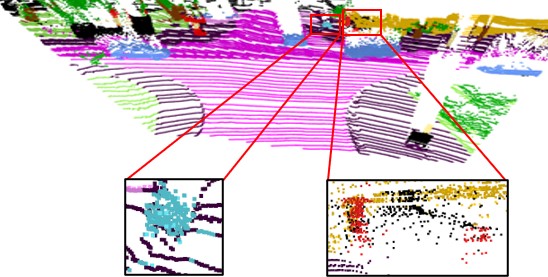}}
\subfigure[Adaptive Filtering]{
\includegraphics[width=0.31\textwidth]{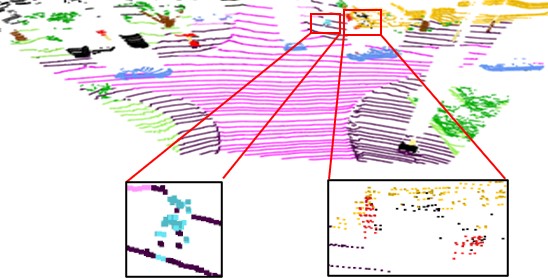}}
\caption{Noise in depth completion and adaptive filtering. (a) Images. (b) Combination of the original point with the depth-completion point (darker in color). (c) Combination of the original point with the adaptively filtered point (darker in color).}
\label{xuni}
\end{figure*}
Due to the density and noise in virtual points generated from RGB images through depth estimation, directly using these points for point cloud semantic segmentation may degrade performance. Specifically, the excessive density of virtual points not only imposes a huge computational burden, but only a minority of these points actually improve performance (see Tab. \ref{tab:6}). Compared to object-wise detection, semantic segmentation is a fine-grained point-wise classification where the introduction of noise points can disrupt the distribution, leading to a decline in performance. Motivated by this observation, we design an adaptive selection approach that leverages voxel density and Euclidean distance to selectively filter and augment the small target point cloud with more reliable virtual points (see Fig. \ref{xuni}).

Under the assumption that pixels around the corresponding pixel are likely to belong to the same class, we map the virtual point $^{V}P$ to RGB pixel coordinates and preliminarily reduce non-essential virtual points by creating a circular region with a radius of two pixels around co-visible points, denoted as ${P}_{v}$. To further filter out more reliable virtual points, we propose a spatial difference-driven adaptive filtering module. Specifically, by voxelizing the initial point cloud $^{L}{P}_{N}$ and the virtual points $ {P}_{v}$ and using the distance $D$ as the boundary, we compute the voxel density on both sides of the circular region, i.e., points-to-voxel grids ratio. We traverse the voxels of the $^{L}{P}_{N}$ and determine the number $N$ of pseudo points to add based on density differences and distance. Then, by the search for the nearest neighbor $N$ of the $^{L}{P}_{N}$ centroid within the voxel, we adaptively select reliable pseudo points from the corresponding virtual voxel grid of ${P}_{v}$. This is described as follows:
\begin{equation}
N=as(\rho + \lceil d/b \rceil )
\end{equation}

where $a$ is the adjustment parameter for the voxel scale, $s$ is the size of the voxel, $\rho$ represents the density difference, $\mathrm{d}=\sqrt{\Bar{x}^{2}+\Bar{y}^{2}+\Bar{z}^{2}}$ is the distance from the centroid to the LiDAR, and $b$ is the adjustment parameter of the distance.

\subsection{Noise-robust Sparse Feature Encoder}
After mitigating point cloud sparsity through adaptive filtering, we propose a sparse encoder that integrates noise-robust feature extraction and fine-grained feature enhancement to extract effective features from noisy merged point clouds. Noise-robust feature extraction reduces noise by encoding geometric features in both 2D image and 3D LiDAR space. Sparse feature enhancement improves feature representation through inner-voxel neighborhood point aggregation and enlarges the receptive field through downsampled voxel aggregation, thereby enhancing sparse geometric features.
\begin{figure}[t]
\centering\includegraphics[width=0.49\textwidth]{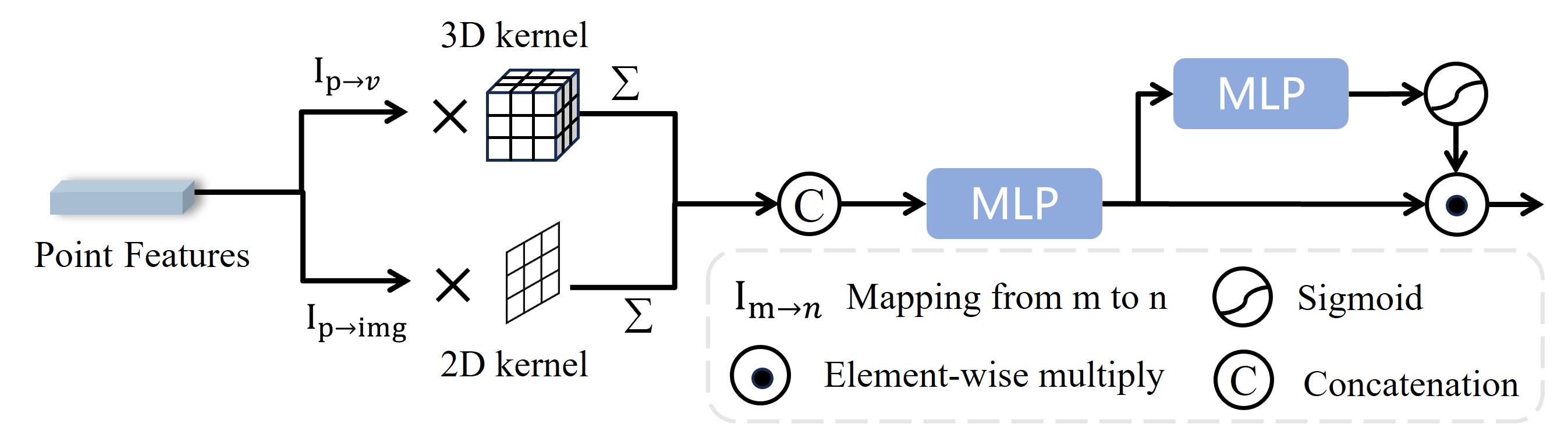}
\caption{The internal structure of noise-robust feature extraction. In each level, we perform sparse submanifold convolutions in both 2D image space and 3D point cloud space and dynamically adjust the feature weights, enhancing feature representation.}
\label{nr}
\end{figure}

$Noise-robust$ $Feature$ $Extraction.$ The virtual points generated by image depth completion networks often contain noise, primarily due to inaccuracies in depth completion. These points are challenging to distinguish in 3D space but are more easily recognized in 2D images \cite{16}. Therefore, we propose a noise-robust feature extraction method, as shown in Fig. \ref{nr}. This module performs sparse submanifold convolutions in both 2D image space and 3D point cloud space to the impact of noise without losing structural cues. Subsequently, a gated attention mechanism dynamically adjusts the feature weights, further enhancing feature representation.

Given the merged point cloud after virtual point adaptive filtering, the initial point characteristics ${F}_{p}\in \mathbb{R}^{N \times{10}}$ consist of the original data $^{L}{P}_{N}$, normalized point clouds within voxel grids, and point clouds offset by voxel centers. In the 3D point cloud space, feature encoding is achieved by scatter $\Phi_{\mathcal{P} \rightarrow \mathcal{V}}^{s}$ to obtain voxel features ${F}_{v}\in \mathbb{R}^{N_{v} \times{C}}$ and performing 3D submanifold convolutions to compute geometric features from non-empty voxels within a 3x3x3 neighborhood as:
\begin{equation}
F^{3D}_{p}=\phi(\mathcal{K}^{3 D}(F_{v})))
\end{equation}
where $\phi(x)$ denotes the nonlinear activation function.

In the 2D image space, feature encoding involves obtaining sparse projection feature maps ${F}_{img}\in \mathbb{R}^{H \times{W} \times{C}}$ through point-to-pixel correspondence and using 2D submanifold convolutions to encode noise-aware features from non-empty features within a 3x3 neighborhood as:
\begin{equation}
F^{2D}_{p}=\phi(\mathcal{K}^{2D}(F_{img})))
\end{equation}
Subsequently, we concatenate these features with MLP encoding to obtain $F^{2D3D}_{p}$, along with a gated attention mechanism, where MLP layers combined with softmax generate dynamic weights, to implicitly learn noise-robust features. These noise-robust features are obtained by:
\begin{equation}
\hat{F}^{2D3D}_{p}=F^{2D3D}_{p} \odot \sigma(\textnormal{MLP}(F^{2D3D}_{p})
\end{equation}
where $\sigma$ denotes the Sigmoid activate function, $\odot$ is element-wise multiply.

\begin{figure}[t]
\centering\includegraphics[width=0.45\textwidth]{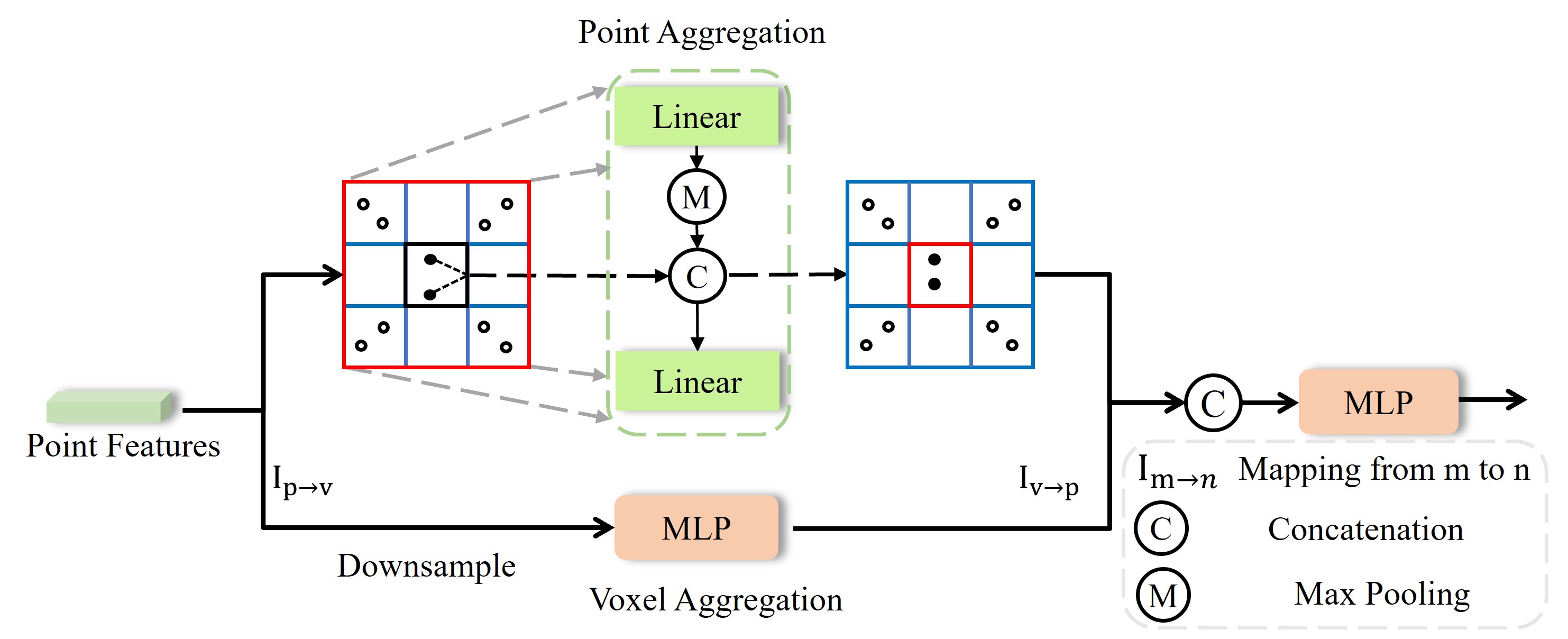}
\caption{Internal structure of the fine-grained feature enhancement (see Algorithm \ref{alg:point_voxel_aggregate_encoder}) which consists of the point feature aggregate and voxel feature aggregation. Point feature aggregation treats the neighboring voxels as points and aggregates point features into each point of the central voxel. Voxel feature aggregation downsamples the voxels at the current scale, extracts features through an MLP and finally indexes the features back to the current point.}
\label{pv}
\end{figure}

\begin{algorithm}[!t]
\caption{Implementation of the Fine-grained Feature Enhancement}
\begin{algorithmic}[1] 
\REQUIRE point feature $F^{2D3D}_{p}$, Scatter $\Phi_{\mathcal{P} \rightarrow \mathcal{V}}^{s}$, ${Gather}$ $\Phi_{\mathcal{V} \rightarrow \mathcal{P}}^{s}$

\STATE $G[V] \leftarrow \text{Apply Scatter} \Phi_{\mathcal{P} \rightarrow \mathcal{V}}^{s}(F^{2D3D}_{p})$

\FOR{each voxel $V$ in $G[V]$}
    \FOR{all points $p$ in $Neighbor(V)$}
        \STATE FC layer and Point feature max-pooling: $F_{\text{max}} \leftarrow \max\{f_c(F^{neighbor}_p), \forall p \in Neighbor(V)\}$
    \ENDFOR
\FOR{each point $p$ in $V$}
    \STATE i) Concatenate $F^{self}_p$ and $F_{\text{max}}$: $F_{\text{concat}} \leftarrow \text{cat}(F^{self}_p, F_{\text{max}})$;
    \STATE ii) Point-wise fully-connected layer: $F_{\text{PA}} \leftarrow f_c(F_{\text{concat}})$
\ENDFOR
\ENDFOR
\STATE Avg downsample: $G_{\text{avg}}[V] \leftarrow \text{Avg Downsample}(G[V])$
\STATE MLP and index to point features: $F_{\text{VA}} \leftarrow \Phi_{\mathcal{V} \rightarrow \mathcal{P}}^{s}(\text{MLP}(G_{\text{avg}}[V]))$
\STATE Concatenate and Point-wise fully-connected layer: $F_{\text{final}} \leftarrow (\text{MLP}f_c(\text{cat}(F_{\text{PA}}, F_{\text{VA}})))$

\RETURN $F_{\text{final}}$
\end{algorithmic}
\label{alg:point_voxel_aggregate_encoder}
\end{algorithm}
$Fine-grained$ $Feature$ $Enhancement.$ After noise-robust feature extraction, a naive way to feature enhancement is point-voxel fusion to extract fine-grained features, as single voxels containing multiple classes of points may produce
ambiguous or wrong predictions \cite{9}. However, existing point-voxel methods face increased computational burden due to additional downsampling and insufficient fine-grained point-level feature aggregation due to the use of global MLP layers. Therefore, we propose the point-voxel geometric feature enhancement method shown in  Fig. \ref{pv}. This method enhances fine-grained feature extraction by aggregating point features within neighboring voxels and expands the receptive field by aggregating downsampled voxels.

Given the noise-robust point features, point feature aggregation maps them to voxels and processes points from surrounding voxels defined by a kernel size of 3, aggregating features from neighboring points for each point within the central voxel, covering a total of 27 voxels. This inner-voxel neighborhood point feature aggregation enhances fine-grained features and effectively effectively eliminates the need for KNN search, streamlining the process to a complexity of $O(1)$ for each point and greatly accelerating the feature aggregation process. The downsampled voxel feature aggregation maps point features to voxels and subsequently downsample the voxel features using $\Phi_{\mathcal{P} \rightarrow \mathcal{V}}^{s}$. By integrating features at both current and downsampled scales with a per-voxel MLP, we increase the effective receptive field and achieve a nuanced, multi-layered feature representation. By concatenating point features obtained through point aggregation with point features from downsampled voxel features through $\Phi_{\mathcal{V} \rightarrow \mathcal{P}}^{s}$, we achieve enhanced sparse geometric features. 
The whole algorithm for the module is illustrated in Algorithm \ref{alg:point_voxel_aggregate_encoder}.

\subsection{Loss Function}
In 3D semantic segmentation, each point is assigned a label. Some methods generate dense feature maps with dense supervision, which leads to severe memory overload. Therefore, we adopt sparse supervision, where voxel semantic labels are generated at different scales during the training phase, and supervision is applied only to active voxels rather than the entire dense feature map. Each voxel label is assigned based on the majority voting of point labels within the voxel. For both main primary and auxiliary supervision, we use the cross-entropy loss and the Lovász-softmax loss \cite{61}. The overall loss of the method is defined by

\begin{equation}\label{e7}
\tilde{L}=\gamma {L}_{c e}+\lambda {L}_{l o v}
\end{equation}

\noindent where $ {L}_{c e}$ and $ {L}_{l o v}$indicate the cross-entropy loss and Lovász-softmax loss, respectively. $\gamma$ and $\lambda$ are the hyperparameters that balance different losses.

\section{Experiments}
We perform experimental evaluations on the large-scale benchmark datasets SemanticKITTI and nuScenes to validate the efficacy of our proposed methods. In addition, we perform an ablation experiment to substantiate the validity of each component proposed in our method.
\begin{table*}[t]
\caption{Comparison on SemanticKITTI validation set. L represents LiDAR-only methods while L+C denotes fusion-based methods. The best and second best scores for each class are highlighted in bold and underline. $^\dagger$ denotes using rotation and translation testing-time augmentations.}
\label{tab:1}
\centering
\resizebox{\textwidth}{!}{%
\begin{tabular}{l|l|lllllllllllllllllll|ll}
\hline
Method  & Input 
& \begin{sideways} {\textcolor{kcar}{$\blacksquare$} car}\end{sideways}
& \begin{sideways} {\textcolor{kbicycle}{$\blacksquare$} bicycle}\end{sideways}
& \begin{sideways} {\textcolor{kmotorcycle}{$\blacksquare$} motorcycle}\end{sideways}
& \begin{sideways} {\textcolor{ktruck}{$\blacksquare$} truck}\end{sideways}
& \begin{sideways} {\textcolor{kothervehicle}{$\blacksquare$} other-vehicle}\end{sideways}
& \begin{sideways} {\textcolor{kperson}{$\blacksquare$} person}\end{sideways}
& \begin{sideways} {\textcolor{kbicyclist}{$\blacksquare$} bicyclist}\end{sideways}
& \begin{sideways} {\textcolor{kmotorcyclist}{$\blacksquare$} motorcyclist}\end{sideways}
& \begin{sideways} {\textcolor{kroad}{$\blacksquare$} road}\end{sideways}
& \begin{sideways} {\textcolor{kparking}{$\blacksquare$} parking}\end{sideways}
& \begin{sideways} {\textcolor{ksidewalk}{$\blacksquare$} sidewalk}\end{sideways}
& \begin{sideways} {\textcolor{kotherground}{$\blacksquare$} other-ground}\end{sideways}
& \begin{sideways} {\textcolor{kbuilding}{$\blacksquare$} building}\end{sideways}
& \begin{sideways} {\textcolor{kfence}{$\blacksquare$} fence}\end{sideways}
& \begin{sideways} {\textcolor{kvegetation}{$\blacksquare$} vegetation}\end{sideways}
& \begin{sideways} {\textcolor{ktrunk}{$\blacksquare$} trunk}\end{sideways}
& \begin{sideways} {\textcolor{kterrain}{$\blacksquare$} terrain}\end{sideways}
& \begin{sideways} {\textcolor{kpole}{$\blacksquare$} pole}\end{sideways}
& \begin{sideways} {\textcolor{ktrafficsign}{$\blacksquare$} traffic-sign}\end{sideways}
& \begin{sideways} mIoU(\%) \end{sideways} & \begin{sideways} speed(ms) \end{sideways}  \\
\hline\hline
Points   (k)          & -     & 6384 & 44      & 52         & 101   & 471           & 127    & 129       & 5            & 21434 & 974     & 8149     & 67           & 6304     & 1691  & 20391      & 882   & 8125    & 317  & 64           & -        & -     \\
\hline
RangeNet++   \cite{35}      & L     & 89.4 & 26.5    & 48.4       & 33.9  & 26.7          & 54.8   & 69.4      & 0.0          & 92.9  &  37.0    &  69.9    & 0.0          & 83.4     & 51.0  & 83.3       & 54.0  & 68.1    & 49.8 & 34.0         & 51.2    & 83.3   \\
SqueezeSegV3 \cite{7}      & L     & 87.1 & {34.3}   & {48.6}       & 47.5  & 47.1          & 58.1   & 53.8      & 0.0          & {95.3}  & 43.1    & 78.2     & 0.3          & 78.9     & {53.2}  & 82.3       & 55.5  & 70.4    & 46.3 & 33.2         & 53.3     & 167   \\
SalsaNext  \cite{36}      & L     & 90.5 &  {44.6}    &  {49.6}       &  {86.3}  &   54.6        &  {74.0}   &  {81.4}      & 0.0          & 93.4  & 40.6    & 69.1     & 0.0          & 84.6     & 53.0  & 83.6       & 64.3  & 64.2    & {54.4} &  {39.8}        & 59.4     & 42   \\
GFNet    \cite{56}    & L     & 94.2 &  49.7    &  62.2     &  74.9  & 32.1     & 69.3   &  83.2      & 0.0          & \underline{95.7}  & 53.8    & $\mathbf{83.8}$     & 0.2     & 91.2    & 62.9  & 88.5      & 66.1  & 76.2    & 64.1 &  48.3    & 63.0     & 100   \\
Cylinder3D   \cite{40}     & L     & 96.4 &  $\mathbf{61.5}$    &  78.2     &  66.3  & 69.8          &  80.8   &  93.3      & 0.0          & 94.9  & 41.5    & 78.0    & \underline{1.4}          & 87.5    & 50.0  & 86.7       & 72.2  & 68.8    & 63.0 &  42.1        & 64.9    & 131    \\
SPVCNN   \cite{53}    & L     & 97.1 &  47.4   &  84.5   &  86.9  & 66.1        &  75.5  &  92.5   & \underline{0.14}    & 93.8  & 52.9 & 81.5     & 0.11 & 91.9  & 65.0  & 89.0   & 68.8  & \underline{77.0}    & $\mathbf{66.8}$ &  50.5      & 67.8     & 259   \\
SphereFormer $^\dagger$ \cite{57}  & L     & 96.9 &  53.7    &  78.3    &  $\mathbf{95.2}$  & 65.4     & 78.4   &  93.6     & 0.0    & 95.1  & 55.1    & 83.1    & $\mathbf{1.8}$     & 91.2   & 61.0  & $ \mathbf{89.5}$     &72.9  & $\mathbf{77.6}$    & \underline{66.5} &  $\mathbf{55.8}$    & 69.0    & $\mathbf{39}$   \\
PointPainting   \cite{12}      & L+C   &  {94.7} & 17.7    & 35.0       & 28.8  &  {55.0}          & {59.4}   & {63.6}      & 0.0          & {95.3}  & 39.9    & 77.6     & 0.4          &  {87.5}     &  {55.1}  & {87.7}       &  {67.0}  & {72.9}    &  {61.8} & 36.5         & 54.5       & - \\
PMF   \cite{13}     & L+C   &  95.4 & 47.8  & 62.9    & 68.4  &  75.2  & 78.9   & 71.6   & 0.0  & $\mathbf{96.4}$ & 43.5   & 80.5     & 0.1   &  88.7  &  60.1 & 88.6       &  72.7  & 75.3  &  65.5 & 43.0    & 63.9      & -  \\
2dpass $^\dagger$  \cite{42} & L+C & \underline{97.8} & 57.6 &  \underline{86.3} &\underline{95.0}  & 82.5 & \underline{83.1} &  \underline{94.5}      & $\mathbf{0.15}$ & 95.1  & $\mathbf{65.9}$    & \underline{83.6}    & 0.05     & $\mathbf{93.5}$   & $\mathbf{73.0} $ & $\mathbf{89.5} $     &\underline{73.4}  & 76.8    &65.6 &  54.6    & $\mathbf{72.0}$      & 62  \\
\hline
VPE(ours)& L+C   & 97.5 & 55.8 & 85.8 & 93.9  &  \underline{83.3}  & 81.4   & 93.3  & \underline{0.14}  & 93.9  & 59.1    & 81.4     & 0.06     & \underline{92.9}     & 69.7  &  88.5   & 71.9  &  75.5  & 62.8 & 53.7   & 70.5    & 70  \\
VPE(ours)$^\dagger$     & L+C   & $\mathbf{98.0}$ & \underline{57.8}    & $\mathbf{87.0}$ & 94.5  & $\mathbf{84.9}$  & $\mathbf{84.1}$   & $\mathbf{94.7}$  & $\mathbf{0.15}$  & 94.9  & \underline{61.3}    & 83.4     & 0.06     & $\mathbf{93.5}$     & \underline{72.6}  &  \underline{89.3}   & $\mathbf{73.5}$  &  76.6  & 65.0 & \underline{54.7}  & \underline{71.8}    & 70  \\
\hline
\end{tabular}%
}
\end{table*}
\subsection{Datasets and Evaluation Metric}
\begin{table*}[t]
\caption{Comparison on nuScenes validation set. L represents LiDAR-only methods, while L+C denotes fusion-based methods. The best and second best scores for each class are highlighted in bold and underline. $^\dagger$ denotes using rotation and translation testing-time augmentations.}
\label{tab:2}
\centering
\resizebox{\textwidth}{!}{%
\begin{tabular}{l|l|llllllllllllllll|ll}
\hline
Method  & Input 
& \begin{sideways} {\textcolor{nbarrier}{$\blacksquare$} barrier}\end{sideways} 
& \rotatebox{90}{\textcolor{nbicycle}{$\blacksquare$} bicycle}	
& \rotatebox{90}{\textcolor{nbus}{$\blacksquare$} bus}
& \rotatebox{90}{\textcolor{ncar}{$\blacksquare$} car}
& \rotatebox{90}{\textcolor{nconstruct}{$\blacksquare$} const. veh.}
& \rotatebox{90}{\textcolor{nmotor}{$\blacksquare$} motorcycle}
& \rotatebox{90}{\textcolor{npedestrian}{$\blacksquare$} pedestrian}
& \rotatebox{90}{\textcolor{ntraffic}{$\blacksquare$} traffic cone}
& \rotatebox{90}{\textcolor{ntrailer}{$\blacksquare$} trailer}
& \rotatebox{90}{\textcolor{ntruck}{$\blacksquare$} truck}
& \rotatebox{90}{\textcolor{ndriveable}{$\blacksquare$} drive. suf.}
& \rotatebox{90}{\textcolor{nother}{$\blacksquare$} other flat}
& \rotatebox{90}{\textcolor{nsidewalk}{$\blacksquare$} sidewalk}
& \rotatebox{90}{\textcolor{nterrain}{$\blacksquare$} terrain}
& \rotatebox{90}{\textcolor{nmanmade}{$\blacksquare$} manmade}
& \rotatebox{90}{\textcolor{nvegetation}{$\blacksquare$} vegetation}
& \begin{sideways} mIoU(\%) \end{sideways} 
& \begin{sideways} speed(ms) \end{sideways}   \\
\hline\hline
Points (k)  &-   & 1629    & 21      & 851  & 6130   & 194          & 81         & 417        & 112          & 370     & 2560  & 56048     & 1972       & 12631    & 13620   & 31667   & 21948      & -   & -      \\ 
\hline
PolarNet  \cite{34} & L    & 74.7    & 28.2    & 85.3 & 90.9   & 35.1         & 77.5       & 71.3       & 58.8         & 57.4    & 76.1  & 96.5      & 71.1       & 74.7     & 74.0    & 87.3    & 85.7       & 71.0   & -   \\ 
Salsanext \cite{36} & L   & 74.8    & 34.1    & 85.9 & 88.4   & 42.2         & 72.4       & 72.2       & 63.1         & 61.3    & 76.5  & 96.0      & 70.8       & 71.2     & 71.5    & 86.7    & 84.4       & 72.2    & -  \\ 
Cylinder3D \cite{40}  & L  & 76.4    & 40.3    & 91.3 & 93.8   & 51.3         & 78.0       & 78.9       & 64.9         & 62.1    & 84.4  & 96.8      & 71.6       &  \underline{76.4}     & 75.4    & 90.5    & 87.4       & 76.1      & 63\\ 
PVKD \cite{58}  & L  & 76.2    & 40.0 & 90.2 & $\mathbf{94.0}$ & 50.9   & 77.4       & 78.8 & 64.9   & 62.0  & 84.1  & 96.6  &  71.4  &  \underline{76.4}     &  \underline{76.3}    & 90.3    & 86.9      & 76.0  & -    \\ 
SPVCNN    \cite{53}  & L  & 76.6 & 48.8 & 92.5 & 86.6 & 54.0   & 86.6 & 80.3 & 66.8 & 62.7  & 84.9  & 96.1  &  72.2  & 73.3  & 74.9  & 87.0 & 86.2  & 76.9   & 63   \\ 
RPVNet \cite{59} & L & 78.2 &  43.4  &  92.7 &  93.2  & 49.0  & 85.7  &  80.5    & 66.0  & 66.9 & 84.0 &  \underline{96.9}  & 73.5  & 75.9  & 76.0 & 90.6 &88.9  & 77.6       & -    \\
OACNN  $^\dagger$ \cite{65} & L & 75.5 &  45.1  &  94.0 &  93.5  & 54.2  & 85.8  &  81.8    & 67.5  & 68.1 & 86.3 & 96.8  & $\mathbf{77.1}$  &  \underline{76.4}  &  $\mathbf{76.6}$ & 89.8 &88.4  & 78.5     & 46      \\
SphereFormer $^\dagger$  \cite{57} & L & 78.7 &  46.7  &  95.2 &  \underline{93.7}  & 54.0 & 88.9   &  81.1  & 68.0   & 74.2  & 86.2    &  $\mathbf{97.2}$   & 74.3 & 76.3   & 75.8  & $\mathbf{91.4}$   &$\mathbf{89.7}$  & 79.5        & $\mathbf{29}$   \\
Ptv3 $^\dagger$  \cite{64} & L & $\mathbf{80.5}$ &  53.9  &  $\mathbf{95.9}$ &  91.9  & 52.3 &  \underline{89.0}   &  84.4  & 71.7   & 74.2  & 84.5    &  $\mathbf{97.2}$    &  \underline{75.6} &  $\mathbf{76.9}$   & 76.1  & \underline{91.2}   & \underline{89.6}  & 80.3      & 45     \\
PMF  \cite{13} & L+C      & 74.1    & 46.6    & 89.8 & 92.1   & 57.0         & 77.7       & 80.9       & 70.9         & 64.6    & 82.9  & 95.5      & 73.3       & 73.6     & 74.8    & 89.4    & 87.7       & 76.9      & 125\\ 
Lif-seg \cite{60}  & L+C      & 76.5    & 51.4    & 91.5 & 89.2   & 58.4         & 86.6       & 82.7       &  \underline{72.9}         & 65.5    & 84.1  & 96.7      & 73.2       & 74.4     & 73.1    & 87.5    & 87.6       & 78.2    & -  \\ 
mseg3D $^\dagger$ \cite{43} & L+C     & 77.6 &   \underline{56.9}    &  \underline{95.8}    &  93.2  & 58.8  & 87.5   &  84.0      & 69.8    & 72.6  & 86.5   & 95.8    & 74.0 & 74.6   & 75.0  & 90.6      &89.3  & 80.1      & 445     \\
2dpass $^\dagger$  \cite{42} & L+C & 78.4 & 52.5    &  95.3    &  93.3  &  $\mathbf{62.4}$     & 88.9   &  83.0      & 68.1 &  \underline{75.6}  & \underline{89.0}    & 96.8    & 75.5    & 76.2   &  75.6  & 89.1      &86.9  & \underline{80.4}     & 44     \\
\hline
VPE(ours)   & L+C   & 77.8 & 53.9    & 95.7  & 91.6  & 63.1          & 88.5  &  \underline{85.2}      & 70.0 &  72.8  & 87.7    & 96.2    & 72.2 & 73.7  & 74.0  & 88.0      & 85.7       & 79.8  & 58    \\
VPE(ours)$^\dagger$     & L+C   & \underline{79.9} & $\mathbf{57.9}$    & $\mathbf{95.9}$  & 93.0  &  \underline{62.2}          & $\mathbf{91.0}$   & $\mathbf{86.9}$      & $\mathbf{73.8}$ &  $\mathbf{76.6}$  & $\mathbf{90.0}$    & 96.8     & 74.5 & 75.9  & 75.6  & 89.1       & 87.0       & $\mathbf{81.6}$    & 58  \\
\hline
\end{tabular}%
}
\end{table*}
In the SemanticKITTI dataset, consisting of 21 sequences, we use only LiDAR data and left-camera images. Sequences 00-10 are semantically annotated, with sequence 08 designated for validation and the rest for training. We also conducted experiments on the NuScenes dataset, which includes 1,000 scenes under various conditions, the data is split into 700 training, 150 validation, and 150 testing scenes, totaling 28,130 training frames and 6,019 validation frames. KITTI \cite{52} has only two front-view cameras, while nuScenes has six cameras that cover the entire field of view.  

To evaluate the performance, we use the Mean Intersection over Union (mIoU) over all classes, which quantifies the overlap between actual and predicted values relative to their combined set. It is calculated by
\begin{equation}
\mathrm{mIoU}=\frac{1}{N} \sum_{c=1}^{N} \operatorname{IoU}_{c}=\frac{1}{N} \sum_{c=1}^{N} \frac{T P_{c}}{T P_{c}+F P_{c}+F N_{c}}
\end{equation}
\noindent where $T P_{c}, F N_{c}, F P_{c}$ denote true positives, false negatives, and false positives, respectively. After averaging the IoU values per class, the primary evaluation metric for semantic segmentation, mIoU, is derived.  
\subsection{Implementation Details}
\begin{table*}[t]
\caption{The class-wise IoU scores of different semantic segmentation approaches on the test set of nuScenes. The best and second best scores for each class are highlighted in bold and underline.}
\label{tab:test}
\centering
\resizebox{\textwidth}{!}{%
\begin{tabular}{l|l|llllllllllllllll|ll}
\hline
Method  & Input 
& \begin{sideways} {\textcolor{nbarrier}{$\blacksquare$} barrier}\end{sideways} 
& \rotatebox{90}{\textcolor{nbicycle}{$\blacksquare$} bicycle}	
& \rotatebox{90}{\textcolor{nbus}{$\blacksquare$} bus}
& \rotatebox{90}{\textcolor{ncar}{$\blacksquare$} car}
& \rotatebox{90}{\textcolor{nconstruct}{$\blacksquare$} const. veh.}
& \rotatebox{90}{\textcolor{nmotor}{$\blacksquare$} motorcycle}
& \rotatebox{90}{\textcolor{npedestrian}{$\blacksquare$} pedestrian}
& \rotatebox{90}{\textcolor{ntraffic}{$\blacksquare$} traffic cone}
& \rotatebox{90}{\textcolor{ntrailer}{$\blacksquare$} trailer}
& \rotatebox{90}{\textcolor{ntruck}{$\blacksquare$} truck}
& \rotatebox{90}{\textcolor{ndriveable}{$\blacksquare$} drive. suf.}
& \rotatebox{90}{\textcolor{nother}{$\blacksquare$} other flat}
& \rotatebox{90}{\textcolor{nsidewalk}{$\blacksquare$} sidewalk}
& \rotatebox{90}{\textcolor{nterrain}{$\blacksquare$} terrain}
& \rotatebox{90}{\textcolor{nmanmade}{$\blacksquare$} manmade}
& \rotatebox{90}{\textcolor{nvegetation}{$\blacksquare$} vegetation}
& \begin{sideways} mIoU(\%) \end{sideways} & \begin{sideways} speed(ms) \end{sideways}   \\
\hline\hline
Points (k)  &-   & 1629    & 21      & 851  & 6130   & 194          & 81         & 417        & 112          & 370     & 2560  & 56048     & 1972       & 12631    & 13620   & 31667   & 21948      & -   & -      \\ 
\hline
PolarNet  \cite{34} & L    & 72.2    & 16.8    & 77.0 & 86.5   & 51.1         & 69.7       & 64.8       & 54.1         & 69.7    & 63.5  & 96.6      & 67.1       & 77.7     & 72.1    & 87.1    & 84.5       & 69.4   & -   \\ 
JS3C-Net \cite{70} & L   & 80.1    & 26.2    & 87.8 & 84.5   & 55.2     & 72.6       & 71.3       & 66.3         & 76.8    & 71.2  & 96.8      & 64.5       & 76.9     & 74.1    & 87.5    & 86.1       & 73.6    & -  \\ 
Cylinder3D \cite{40}  & L  & 82.8    & 29.8    & 84.3 & 89.4   & 63.0         & 79.3       & 77.2       & 73.4    & 84.6    & 69.1  &$\mathbf{97.7}$     & \underline{70.2}       & 80.3     & 75.5    & 90.5    & 87.6       & 77.2      & 63\\ 
AMVNet   \cite{37} & L  & 80.6    & 32.0 & 81.7 & 88.9 & 67.1   & 84.3   & 76.1 & 73.5   & 84.9  & 67.3 & 97.5  &  67.4  & 79.4     & 75.5    & 91.5    & 88.7      & 77.3  & -    \\  
$(AF)^{2}$-S3Net \cite{41} & L & 78.9 &  52.2  &  89.9 &  84.2  & \underline{77.4}  & 74.3  &  77.3    & 72.0  & 83.9 & 73.8 & 97.1  & 66.5  & 77.5  & 74.0 & 87.7 &86.8  & 78.3       & 270    \\
RangeFormer \cite{71} & L &  $\mathbf{85.6}$ &  47.4  &  91.2 &  90.9  & 70.7  & \underline{84.7}  &  77.1    & 74.1  & 83.2 & 72.6 & 97.5  & $\mathbf{70.7}$  & 79.2  &  75.4 & 91.3 &88.9  & 80.1     & -      \\
PMF  \cite{13} & L+C      & 82.0   & 40.0    & 81.0 & 88.0   & 64.0         & 79.0       & 80.0       & 76.0     & 81.0    & 67.0  & 97.0      & 68.0       & 78.0     & 74.0    & 90.0    & 88.0       & 77.0      & 125\\  
2D3DNet \cite{44}  & L+C & 83.0 & \underline{59.4}  &  88.0 &  85.1  & 63.7  & 84.4  &  82.0    & 76.0  & 84.8 & 71.9 & 96.9  & 67.4  & 79.8  &  76.0 & \underline{92.1} &\underline{89.2}  & 80.0     & -      \\
2dpass   \cite{42} & L+C & 81.7 & 55.3    &  \underline{92.0}    &  \underline{91.8}  & 73.3     &  $\mathbf{86.5}$   &  78.5      & 72.5 & 84.7 & 75.5    & 97.6    & 69.1    & 79.9   & 75.5  & 90.2      &88.0  & 80.8     & $\mathbf{44}$     \\
mseg3D \cite{43} & L+C     & 83.1 &  42.4    & $\mathbf{94.9}$    &  91.0  & 67.1  & 78.5   & $\mathbf{85.6}$      &$\mathbf{80.4}$  & $\mathbf{87.5}$   &$\mathbf{77.3}$    &$\mathbf{97.7}$    & 69.8 & $\mathbf{81.2}$   &$\mathbf{77.8}$  &$\mathbf{92.3}$       &$\mathbf{90.0}$  & \underline{81.1}     & 445     \\
\hline
Baseline  & L  & 80.0 & 30.0 & 91.9 & 90.8 & 64.7   & 79.0 & 75.6 & 70.9 & 81.0  & 74.6  & 97.4  &  69.2  & 80.0  & 76.1  & 89.3 & 87.1  & 77.4   & 63   \\
VPE(ours)   & L+C   & \underline{84.2} & $\mathbf{63.2}$    & 91.0  & $\mathbf{92.1}$  &  $\mathbf{79.7}$  & 82.4   & \underline{83.8}      & \underline{77.8}  &  \underline{86.5}  & \underline{75.6}    & \underline{97.6}     & \underline{70.2} & \underline{80.8}  &\underline{76.7}  & \underline{92.1}       & $\mathbf{90.0}$       & $\mathbf{82.7}$    & 58  \\
\hline
\end{tabular}%
}
\end{table*}
We choose an extended SPVCNN \cite{53} encoder as our baseline 3D network. This structure has an initial voxel size of 0.1, fewer parameters, and a hidden dimension of 256. For multi-scale feature layers, we use a downsampling stride of [2, 4, 8, 16, 16, 16]. Meanwhile, we use random point dropout, random flipping, and random scaling during the training stage. We train our model using a single GeForce RTX 4090 GPU. The training uses an SGD optimizer with a cosineannealingLR schedule for 80 epochs. The momentum is set at 0.9, the learning rate is set at 0.24, and the weight decay at 0.0001. We preprocess the input scene by limiting it to the ranges from [-51.2m, -51.2m, -4m] to [51.2m, 51.2m, 2m] on SemanticKITTI and [-50m, -50m, -4m] to [50m, 50m, 3m] on nuScenes. The depth estimation network is PENet \cite{54} on SemanticKITTI and MVP \cite{45} on nuScenes. Additionally, in the spatial difference-driven adaptive filtering module, the distance interval $D$ is set to 20 and the voxel grid $s$ is set to 0.2 for the SemanticKITTI dataset. For the nuScenes dataset, the $D$ is set to 10 and the $s$ is set to 0.4. The voxel ratio adjustment parameter $a$ is set to 5, and the distance adjustment parameter $b$ is set to 20. Test-time augmentation is also applied during inference.
\subsection{Performance Results and Analyses}
\begin{figure*}[t!]
\centering
\includegraphics[width=1.0\textwidth]{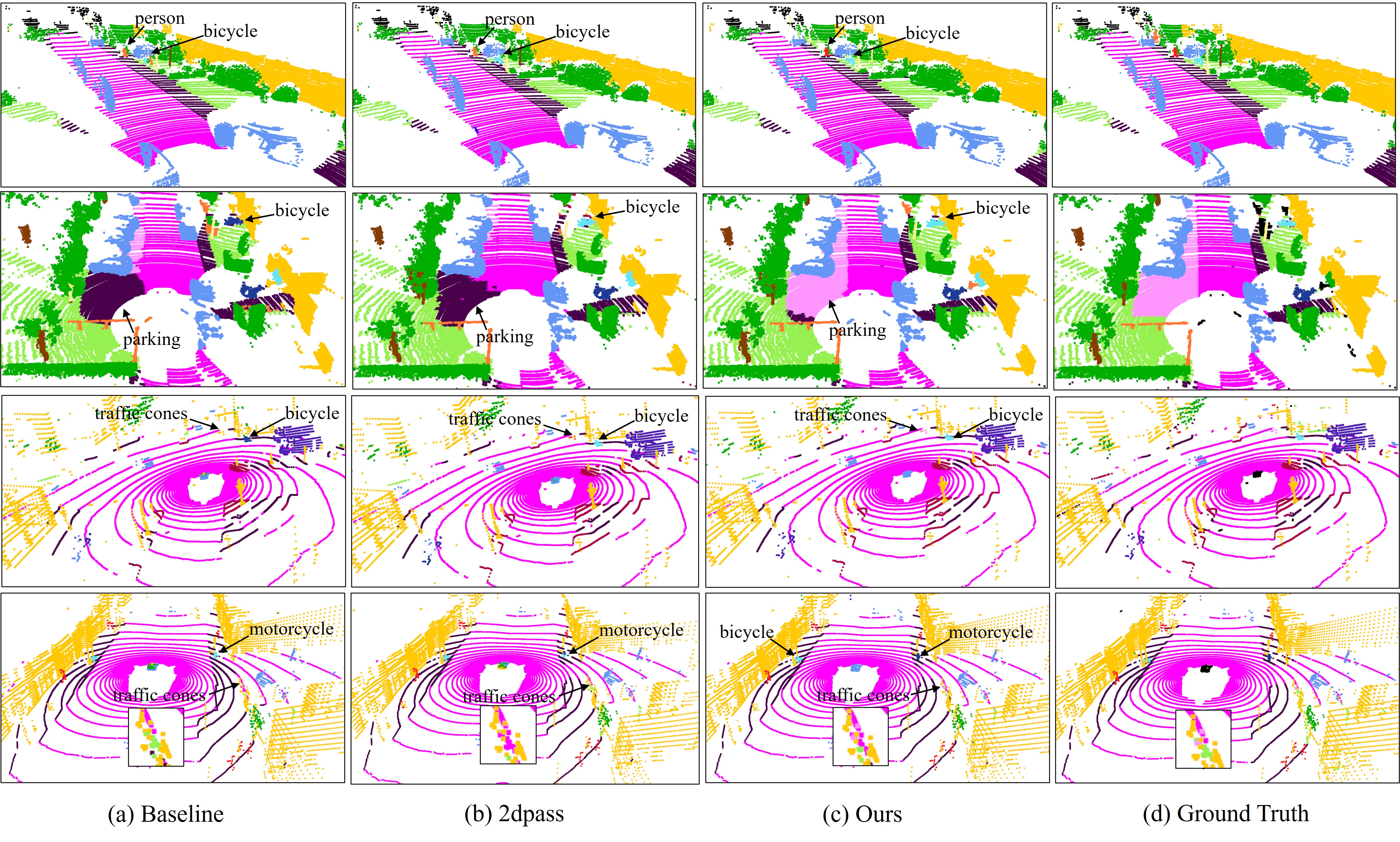}
\caption{Qualitative results of VPE on the validation set of SemanticKITTI and nuScenes. Each pair of rows displays results from SemanticKITTI and nuScenes, respectively. Within each row, images from left to right represent the baseline, 2dpass, our results, and the ground truth.}
\label{keshihua}
\end{figure*}
$Results$ $on$ $nuScenes.$ In Tab. \ref{tab:test} and Tab. \ref{tab:2}, we show the results of the semantic segmentation in the nuScenes test and val set, respectively. It can be observed that research on multi-modal methods is relatively less frequent compared to methods using only LiDAR. Compared to previous methods, our proposed framework exhibits the best performance in mIoU, demonstrating its excellence. As shown in the Tab. \ref{tab:2}, our method achieves superior performance on challenging small objects such as bicycles, pedestrians, and traffic cones. This improvement is primarily driven by the introduction of virtual point enhancement, effectively mitigating the limitations posed by insufficient point density that typically hampers segmentation in LiDAR-only approaches. The visualization on nuScenes validation set is shown in Fig.\ref{keshihua}. Compared to the baseline, our method successfully identifies small targets, such as traffic cones and accurately segments bicycles that were misclassified by the baseline.

$Results$ $on$ $SemanticKITTI.$ The Tab. \ref{tab:1} shows the experimental comparison on SemanticKITTI. Compared to single LiDAR methods, our approach achieves the best performance. However, compared to multi-modal methods, it falls slightly short because the 64-line Semantickitti dataset is already dense enough, resulting in a less significant performance improvement from virtual point enhancement compared to the 32-line nuScenes dataset. The visualizations on the SemanticKITTI validation set are shown in Fig. \ref{keshihua}. 
\begin{figure}[t]
\centering\includegraphics[width=0.4\textwidth]{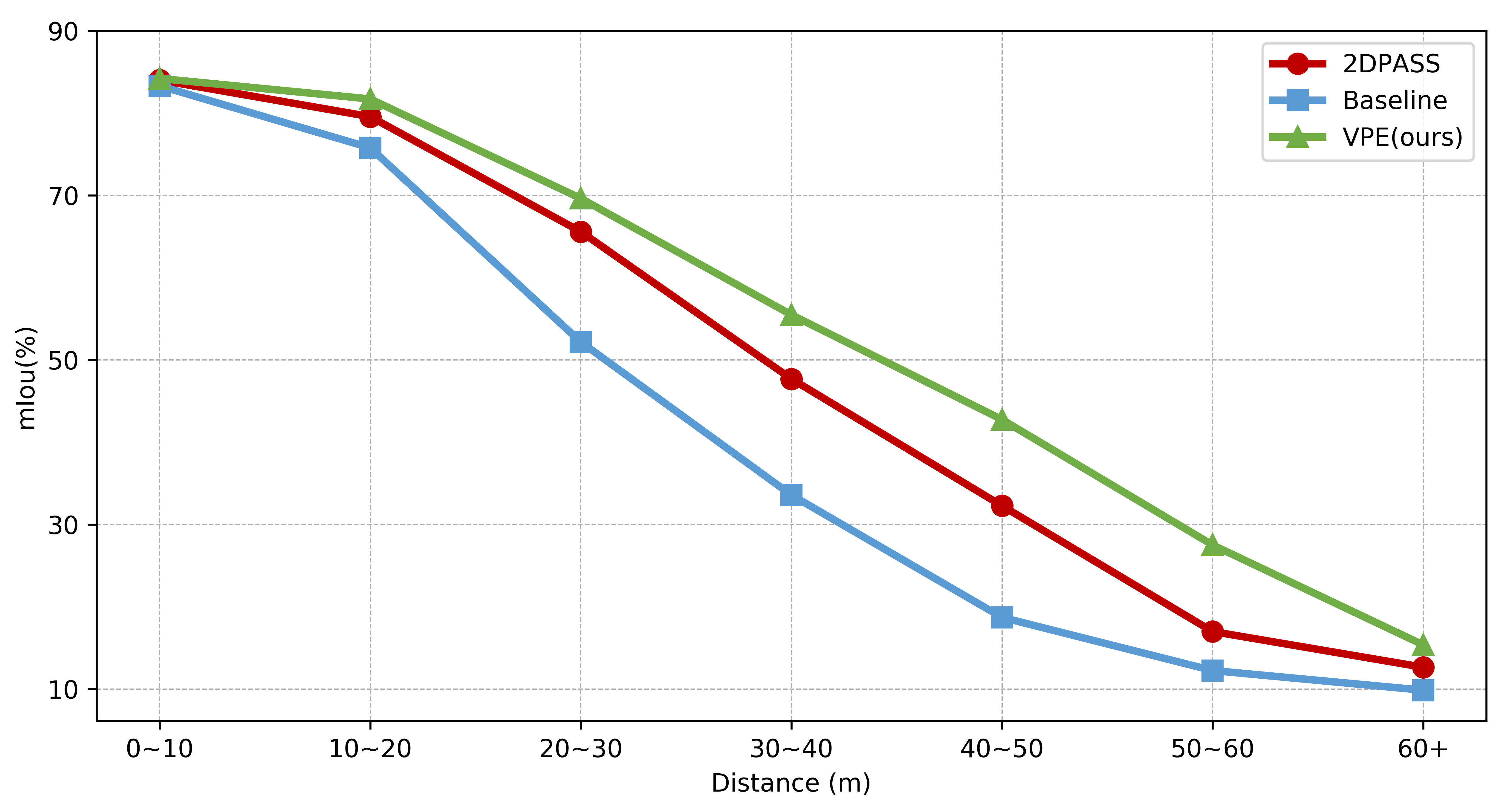}
\caption{ Distance-based evaluation on nuScenes. As the distance increases, the point cloud becomes sparse.}
\label{dis}
\end{figure}

$Comparing$ $with$ $other$ $point-voxel$ $methods.$ To further validate the effectiveness of our proposed point-voxel module, we compared VPE with typical point-voxel methods listed in Tab \ref{tab:7}, using these methods in the same single sensor to ensure a fair comparison. Low-resolution voxelization struggles to accurately recognize small instances, while the point-voxel module effectively addresses this issue by introducing a high-resolution point branch. We performed a comparative analysis based on mIoU and inference time. As shown in the table, VPE demonstrates superior performance in mIoU and maintains a competitive inference time.
\begin{table}[t]
\centering
\caption{Comparison with different point-voxel methods in single model on the nuScenes.}
\label{tab:7}
\resizebox{0.75\columnwidth}{!}{%
\begin{tabular}{c|c|c}
\hline
Method  & mIoU(\%) &speed (ms) \\
\hline
PCSCNet \cite{66}  & 72.0\% & \textbf{44} \\
SPVCNN \cite{53} & 76.9\%   & 63 \\
CPGNet \cite{67}  & 76.9\%   & 60  \\
RPV-CASNet \cite{68}& 77.1 \%  & 82 \\
VPE(ours)  & \textbf{79.2\%}  & 48 \\
 \hline
\end{tabular}
}
\end{table}
\begin{figure}[t]
\centering\includegraphics[width=0.4\textwidth]{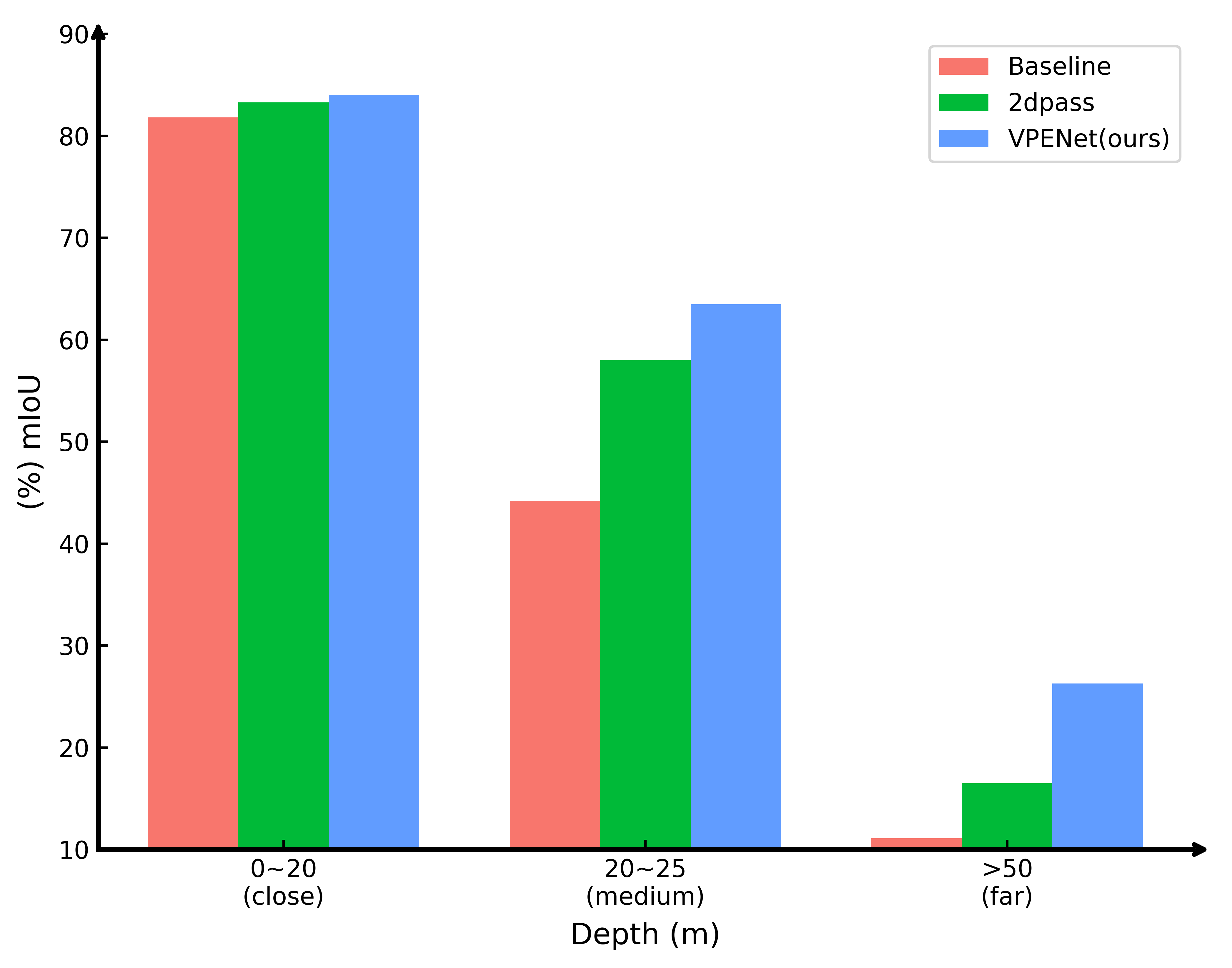}
\caption{ Performance evaluation at close, medium, and long ranges on the nuScenes. Our method demonstrates significantly greater improvement at medium and long ranges compared to close ranges.}
\label{dis_mid}
\end{figure}

$Distance-based$ $Evaluation.$ We investigate how the distance from the ego vehicle affects the segmentation performance and compare it with VPE, the current state-of-the-art techniques, and the baseline on the nuScenes validation dataset. Fig. \ref{dis} clearly illustrates VPE's mIoU relative to the baseline and 2DPASS. As the distance increases, all methods show a gradual decline in results due to lower point density at longer distances. VPE significantly improves performance at medium to long distances, due to virtual point enhancement. Conversely, the baseline, which relies on a single LiDAR, experiences a steep decline in performance with increasing distance. The multi-modal fusion of 2DPASS, which incorporates camera information, helps mitigate this performance degradation trend. To evaluate the performance improvements of our proposed method at medium and long distances, we categorize them as close (less than $20m$), medium (between $20m$ and $50m$), and far (greater than $50m$). As shown in Fig. \ref{dis_mid}, our method demonstrates substantially greater improvement in medium and long ranges compared to close ranges.
\begin{table}[t]
\centering
\caption{Comparison of co-visible point performance improvement on the nuScenes dataset.}
\label{tab:8}
\resizebox{0.7\columnwidth}{!}{%
\begin{tabular}{c|c}
\hline
Method  & Co-points mIoU(\%)  \\
\hline
Baseline  & 76.1\%  \\
2dpass & 79.9\%   \\
VPE(ours)  & \textbf{81.3\%}  \\
 \hline
\end{tabular}
}
\end{table}
$Co-points$ $Evaluation.$ Existing methods are predominantly LiDAR-based. We also evaluate the performance improvements in co-visible points achieved by multimodal approaches compared to single LiDAR methods. As shown in the Tab. \ref{tab:8}, our method significantly outperforms the SOTA method 2DPass in improving the performance on co-visible points.
\subsection{Ablation Study}
$Component$ $Analysis.$ To validate the effectiveness of each module, we conduct an extensive ablation study and list the results on the nuScenes validation dataset in Tab. \ref{tab:3}. As shown in Tab. \ref{tab:3}, our baseline achieved a relatively modest mIoU of 76.89. The introduction of virtual point augmentation and the spatial difference-driven adaptive filtering module increased the mIoU to 79.78 (virtual points constitute merely 9.6\% of the total point cloud). By integrating the sparse encoder that encompasses noise-robust feature extraction and fine-grained feature enhancement, the mIoU improved to 81.66. Additionally, we also explored the performance improvement of each module on the mIoU of co-visible points. 
\begin{table}[t]
\caption{Ablation study on the nuScenes validation set. SDAFM: Spatial Difference-Driven Adaptive filtering module. NSFE: Noise-robust Sparse feature encoder. NFE: Noise-robust feature extraction. FFE: Fine-grained feature enhancement. Metric: mIoU.}
\label{tab:3}
\centering
\resizebox{0.7\columnwidth}{!}{%
\begin{tabular}{c|c|cc|c}
\hline \multirow{2}{*}{\text { Baseline }} & \multirow{2}{*}{\text {SDAFM}}  & \multicolumn{2}{c|}{\text {NSFE}}  & \multirow{2}{*}{\text{ mIoU$\uparrow$}} \\
& & \text {NFE} & \text {FFE} & \\
\hline 
\text{\checkmark} & & & & 76.89  \\
\text{\checkmark} & \text{\checkmark} & & & 79.78 \\
\text{\checkmark} & & \text{\checkmark} & & 79.24 \\
\text{\checkmark}  & & \text{\checkmark} & \text{\checkmark} & 80.71  \\
\text{\checkmark} & \text{\checkmark} & \text{\checkmark} & \text{\checkmark} & 81.66 \\
\hline
\end{tabular}%
}
\end{table}
\begin{table}[t]
\centering
\caption{Ablation study of the hyperparameter used in spatial difference-driven adaptive filtering module. Percentage indicates the proportion of virtual points in the total point cloud.}
\label{tab:6}
\resizebox{0.7\columnwidth}{!}{%
\begin{tabular}{cc|c|c}
\hline
$s$ & $d$ & mIoU$\uparrow$  & Percentage \\
\hline
 - & - & -0.39  & 22.80\% \\
 0.2 & 10 & 2.36  & 5.21\% \\
0.4 & 10 & \textbf{2.89}  & 7.73\% \\
0.2 & 20 & 1.25 & 2.05\% \\
0.4 & 20 & 1.87  & 3.20\% \\
 \hline
\end{tabular}
}
\end{table}

$Adaptive$ $Filtering.$ In addition, the voxel grid $s$ and the distance threshold $D$ in the spatial difference-driven adaptive filtering module control the scale of valid virtual points. We vary these hyperparameters and show the results in Tab. \ref{tab:6}. The results indicate that unfiltered virtual points account for 22\% of the total point cloud, increasing the computational load and decreasing mIoU performance. In contrast, after adaptive filtering, introducing fewer virtual points effectively improves the mIoU metric. Specifically, with a voxel grid of 0.4 and a distance threshold of 10, introducing 7.73\% of virtual points resulted in a 2.89\% improvement in mIoU performance.

\section{Conclusions}
In this paper, we introduce a multi-modal semantic segmentation approach based on virtual point enhancement (VPE), which improves the performance of LiDAR semantic segmentation by enhancing the sparsity of point clouds with dense but noisy virtual points. VPE employs a spatial difference-driven adaptive filtering module to obtain reliable pseudo points, thereby mitigating the point cloud density for medium-range and small targets. Furthermore, we propose a noise-robust sparse feature encoder that incorporates noise-robust feature extraction and fine-grained feature enhancement. Noise-robust feature extraction exploits the 2D image space to reduce the impact of noisy points, while fine-grained feature enhancement boosts sparse geometric features through inner-voxel neighborhood point aggregation and downsampled voxel aggregation. The results on SemanticKITTI and nuScenes, two large-scale benchmark datasets, have validated effectiveness, significantly improving 2.89\% mIoU with the introduction of 7.7\% virtual points on nuScenes. We believe that our work has the potential for broader applications in autonomous driving, such as in tasks like 3D reconstruction and semantic completion.

\section*{Acknowledgments}
This work was supported by the National Natural Science Foundation of 61991412. 



\bibliographystyle{IEEEtran}
\bibliography{mybib}

\end{document}